\documentclass{article}
\usepackage{ijcai24}

\usepackage{times}
\usepackage{soul}
\usepackage{url}
\usepackage[hidelinks]{hyperref}
\usepackage[utf8]{inputenc}
\usepackage[small]{caption}
\usepackage{graphicx}
\usepackage{amsmath}
\usepackage{booktabs}
\usepackage{algorithm}
\usepackage{algorithmic}
\usepackage[switch]{lineno}

\usepackage{amssymb}
\usepackage{xcolor}
\usepackage{subfigure}

\usepackage{etoolbox,lipsum}

\usepackage[thmmarks,amsmath]{ntheorem}
\theoremnumbering{arabic}
\theoremstyle{break}
\theoremsymbol{}
\theoremheaderfont{\normalfont\bfseries}
\theorembodyfont{\upshape}

\title{CompetEvo: Towards Morphological Evolution from Competition}

\author{
Kangyao Huang$^1$\and
Di Guo$^2$\and
Xinyu Zhang$^{3}$\and
Xiangyang Ji$^{4}$\And
Huaping Liu$^1$\footnote{Huaping Liu is the corresponding author.}
\affiliations
$^1$Department of Computer Science and Technology, Tsinghua University\\
$^2$School of Artificial Intelligence, Beijing
University of Posts and Telecommunications\\
$^3$School of Vehicle and Mobility, Tsinghua University\\
$^4$Department of Automation, Tsinghua University\\
\emails
huangky22@mails.tsinghua.edu.cn,
guodi.gd@gmail.com,
\{xyzhang, xyji, hpliu\}@tsinghua.edu.cn
}

\begin{document}

\maketitle

\begin{abstract}
Training an agent to adapt to specific tasks through co-optimization of morphology and control has widely attracted attention. However, whether there exists an optimal configuration and tactics for agents in a multiagent competition scenario is still an issue that is challenging to definitively conclude. In this context, we propose competitive evolution (CompetEvo), which co-evolves agents' designs and tactics in confrontation. We build arenas consisting of three animals and their evolved derivatives, placing agents with different morphologies in direct competition with each other. The results reveal that our method enables agents to evolve a more suitable design and strategy for fighting compared to fixed-morph agents, allowing them to obtain advantages in combat scenarios. Moreover, we demonstrate the amazing and impressive behaviors that emerge when confrontations are conducted under asymmetrical morphs.
\end{abstract}

\section{Introduction}

The auto-generation of agent design has been studied in embodied intelligence currently. Building an effective agent entails adjusting its physical design and actions for the environment and tasks, posing the co-design challenge of morphology and controller~\cite{Chen,liu2023morphology}. This concept is also known as body-brain co-optimization, morph-control co-evolving, or design-control co-optimization.

Previous studies aim to design agent morphologies that are better suited for environments, limited to the scenarios of one-player simple tasks like moving and jumping~\cite{Yuan2022,Wang2019,Ha2019,Sims2023,Chen,Cai,Wang2023,Zhang2022f,Zhang2022g}. 
On the other hand, achieving evolution through cooperation and competition among multiple individuals is pervasive in the biological realm\cite{Huang2024,Huang2021c,Huang2022c,Wang2023c,Chen2023b}, especially on the morphological aspect. For example, athletes usually undergo extensive training to develop and reshape their bodies, along with honing their skills, to achieve better performance in formal competitions. Inspired by this, we redirect our attention to the co-evolution of morph and combat tactics within competitions.

In this study, our emphasis is on the strategy that co-evolves morph and tactics within two-player games, and to the best of our knowledge, it is the first endeavor to incorporate embodied morphological evolution into adversarial games. Moreover, this opens avenues to explore how the physical attributes of agents can be optimized for confrontations. The main contributions are:
\begin{itemize}
    \item We propose \textbf{CompetEvo} that co-evolves agent morphology and fighting tactics to integrate embodied morphological evolution into competitive games.
    \item A series of cross-antagonism experiments is conducted to validate the significant role played by morphological evolution during confrontations in enhancing an agent's ability to deal with adversaries.
    \item We showcase remarkable emergent behaviors demonstrated by agents when morphological evolution is permitted during competition, surpassing our initial expectations.
\end{itemize}

This paper is organized as follows: Section~\ref{related works} provides an overview of existing relevant work and methods; Section~\ref{problem definition} defines the problem; in Section~\ref{methodology} and Section~\ref{training curriculum}, our proposed method and training approach are described in detail, then the effectiveness and some interesting cases of our method are illustrated in Section~\ref{experiments}; Section~\ref{conclusion} makes a conclusion.

\begin{figure*}[t]
    \includegraphics[width=\textwidth]{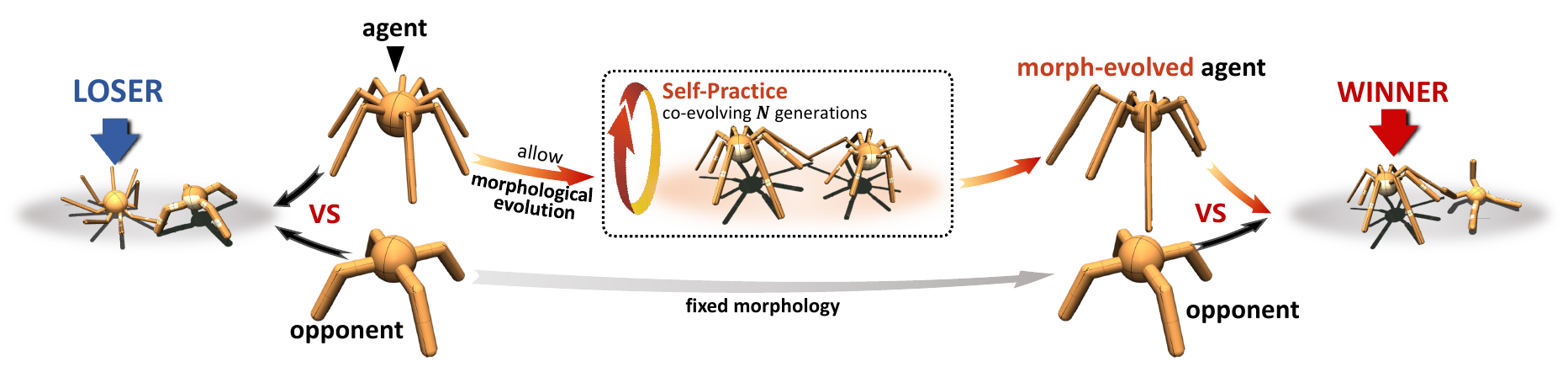}
    \caption{The key insight of this article: agent in its original morph is at a disadvantage in competitive confrontations with the opponent. However, after undergoing \textit{\textbf{N}} generations of co-evolution in both morphology and tactics, agent with new morphology and combat tactics can overcome the original opponent in competition. Using \texttt{spider} and \texttt{ant} as an example.}
    \label{fig:competevo}
\end{figure*}

\section{Related Works}
\label{related works}
\subsection{Two-player Games in Embodied AI}
Artificial Intelligence (AI) has shown its advantages in two-player games like Go and Chess~\cite{Silver2016,Silver2018}. When considering embodied AI in two-player games, OpenAI provides amazing results that promote the simulation of agents at the control and execution levels~\cite{Bansal2018a}. For two-player games where agents have asymmetric morph, meta-learning methods are proposed to solve a long-term continuous adaptation~\cite{MaruanAl-ShedivatTrapitBansalYuraBurdaIlyaSutskeverIgorMordatch2013}. More embodied learning platforms of multiagent cooperation and competition are also proposed by DeepMind~\cite{Liu2022,Liu2019,Haarnoja2023}. However, these studies are based on agents with fixed morphs, without considering whether the body morphology of the agents is truly suitable for the tasks.

\subsection{Co-evolution of Agent Morph and Control}
Agents with morphologies suited for the task often have a greater likelihood of gaining better performance in the given task, and improving task performance through the adjustment of an agent's morphology and control has been a widely studied and long-standing issue~\cite{Chen}.

There are two mainstreams based on whether structural topology changes or not. One category of studies optimizes agent attributes, functionalities, and design parameters~\cite{Schaff2019,Chen,Ha2019}; others focus more on optimizing structure based on the diagram description~\cite{Wang2019,Yuan2022,Hu2023} by graph neural network (GNN)~\cite{Wang2018}.

Population-based methods are the most general co-evolution frameworks, which maintain a bi-level optimization, and regard morph and control as the outer and inner objectives, respectively~\cite{Ha2019}. Evolutionary searching (ES) is also used ~\cite{Gupta2021,Wang2019} in outer optimization to generate embodied agents. On the other hand, some researchers attempt to solve a multi-objective joint optimization problem directly~\cite{Yuan2022}.

Before this work, researchers have explored training agents with diverse morphologies and functionalities, fostering collaboration or competition in StarCraft \uppercase\expandafter{\romannumeral2}~\cite{Yuan2023,Vinyals2019a}. Nonetheless, these endeavors did not focus on training evolvable morphologies and functionalities, marking the primary distinction from our research efforts.

\section{Problem Definition}
\label{problem definition}
We first employ a two-player Markov game~\cite{Gleave2020} to formalize the problem, defined by: a set of states $\mathcal{S}$ describing the state of the world and possible states of both players, a set of actions of each player $\mathcal{A}_{\alpha}$ and $\mathcal{A}_{\beta}$ where we distinguish the agent and opponent by subscripts $\alpha$ and $\beta$, a joint transition function $\mathcal{T}:\mathcal{S}\times \mathcal{A}_{\alpha } \times \mathcal{A}_{\beta}\rightarrow \mathcal{S} $ determining distribution over the next states, reward functions $R_{i}$ $:\mathcal{S}\times \mathcal{A}_{\alpha} \times \mathcal{A}_{\beta} \rightarrow \mathbb{R}, i \in \{ \alpha, \beta \}$, contingent upon the current states and actions generated by each agent.

Previous work has studied two-player Markov games~\cite{Bansal2018a} and developed some interesting fighting skills. Based on this, we introduce morphological evolution, expanding the problem from the sole optimization of adversarial strategies to the joint optimization of confrontational morphology and fighting tactics, aiming to acquire improved morphologies tailored specifically for adversarial games. This promotion is highly challenging because it is not only evident in the morphology and skeletal structure of the agents but also manifested in their adept utilization of the evolved body to reinforce their combat skills, giving rise to novel fighting tactics.

To introduce morphological evolution, we formulate $M_{i}$ to delineate the morph player $i\in \{\alpha, \beta\}$, which describes agent configurations like bone length and size, as well as joint limitations, and is involved in state sets $\mathcal{S}$. Here we define the co-evolution combined policy as $ \pi$ that includes morph sub-policy with parameter $\theta$ and fighting tactics sub-policy with parameter $\phi$. Given the predefined adversarial task and its corresponding rules, each player $i \in \{ \alpha, \beta \}$ refines its policy $\pi_{i}(\theta;\phi)$, which generates a physical agent morphology and provides tactics for fighting. At the beginning of every game, evolvable players prepare an evolved morph $M_{i}$ generated by morph sub-policy for combat. Each player aims to maximize its total expected return over time horizon $T$ under confrontation. The cost function for agent $i$ under morph $M_{i}$ can be described as $J(\pi_{i}, M_i) = \mathbb{E}_{\pi_{i}, M_{i}}\left [\sum_{t=0}^{T} \gamma^t R_{i} \right ]$, where $\gamma$ is the discount factor for rewards, and $T$ denotes the time horizon. A special case is when neither agent changes their morphs, this problem degrades to a classical two-player game studied in ~\cite{Bansal2018a,MaruanAl-ShedivatTrapitBansalYuraBurdaIlyaSutskeverIgorMordatch2013,Yuan2023}.

\section{Methodology}
\label{methodology}
The considered fighting scenario is illustrated in Figure~\ref{fig:competevo}. Initially, the agent with the original morph cannot defeat the opponent. However, by permitting optimization on morphology and co-evolving the agent through self-practice, the resulting morph-evolved agent has the potential to dominate the same opponent in combat. This two-player game faces some serious challenges: how to maintain a self-practice two-player training continuously and how to optimize agents' morph and tactics jointly. In this section, we introduce the details about the proposed method.

\subsection{Continuous Self-Practice Training}

In the training of two-player games, a crucial aspect is to avoid significant imbalances in confrontations, as they may lead to divergent training strategies~\cite{Bansal2018a}: the disadvantaged agent struggles to acquire useful information in the competition while the opponent, although in an advantageous position, lacks robustness in its strategies. Choosing appropriate opponents can address this issue. Our objective is to ensure that the agents' capabilities are continuously and collaboratively enhanced during confrontations. Therefore, the choice of an opponent becomes a problem that affects the stability of training, especially since the morphs of agents are also changeable in our environment.

To overcome this and refer to other two-player competition tasks, our training utilizes $\delta-Uniform$ opponent sampling, which ensures continual learning by training a policy that could consistently beat random previous versions of the opponent. Thus, the agent of the current policy from $\mathcal{P}_{\alpha}$ separately plays with multiple opponents of the previous policies from set $\mathcal{P}_{\beta}$. Let $\delta \in [0, 1]$ represent the percentage threshold applied to the oldest policy, determining its eligibility for potential sampling from the opponent player pool. At the commencement of each episode, we always choose the agent of the current policy and uniformly sample $\mathcal{N}$ historical opponent policies, depending on the specified $\delta$ threshold.

\subsection{Morph Evolution and Tactics}
We introduce the co-evolution strategy into confrontation. First, we demonstrate how we encode agent morphology into parameters. Subsequently, we provide details of morphology optimization in adversarial games.

\subsubsection{Morphology Encoding}
\begin{figure}
    \centering
    \includegraphics[width=\columnwidth]{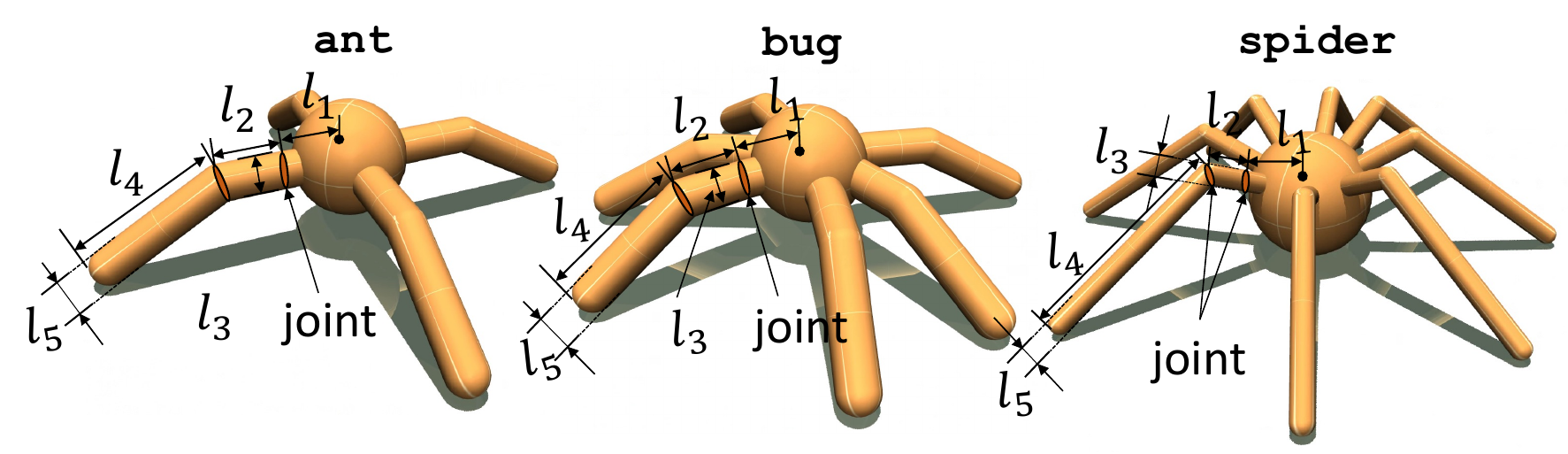}
    \caption{Morphology encodings of three different agents: \texttt{ant}, \texttt{bug}, and \texttt{spider}. The encoding methods are specific to the legs. We define 20, 30, and 40 parameters to describe their designs, respectively.}
    \label{fig:design encoding}
\end{figure}

We mainly use three types of species and their evolved derivatives. \texttt{ant}, \texttt{bug}, and \texttt{spider}, as shown in Figure~\ref{fig:design encoding}; then, based on these original morphs, we develop their evolvable versions: \texttt{evo-ant}, \texttt{evo-bug}, and \texttt{evo-spider}. The evolvable versions can adjust the morphological parameters of legs and the action capabilities corresponding to their morphs.

The agent morphology with fixed topology can be naturally represented using predefined variables~\cite{Ha2019}. We encode the designs of three animals using structural rules, illustrated in Figure~\ref{fig:design encoding}. All legs of animals share a similar structure: two joints and two limbs. The difference lies in that each species has a different number of legs: four for \texttt{ant}, six for \texttt{bug}, and eight for \texttt{spider}. Therefore, we use a vector with five parameters to describe one leg. In detail, $l_2, l_3, l_4, l_5$ denote the length and size of the thigh and lower leg respectively, and $l_1$ indicates the distance from thigh to torso. To build links between morph and capability, the limb size maintains a linear relationship with the joint ability (moment arms, velocity and force). In brief, the more robust limb can generate greater power. Subsequently, we encode agent design $M$ into a vector $m=(l_1, ..., l_n)$. It is worth mentioning that variables in $m$ are the scaling factor relative to the original agent but not the actual physical length. This is similar to a normalization of the real parameters, which makes morph optimization easier. Moreover, we limit the scaling of limbs to avoid a great disparity in size because our arenas have limited size and excessively large bodies may exceed the spatial constraints of the environment.

\begin{figure}[t]
    \centering
    \includegraphics[width=\columnwidth]{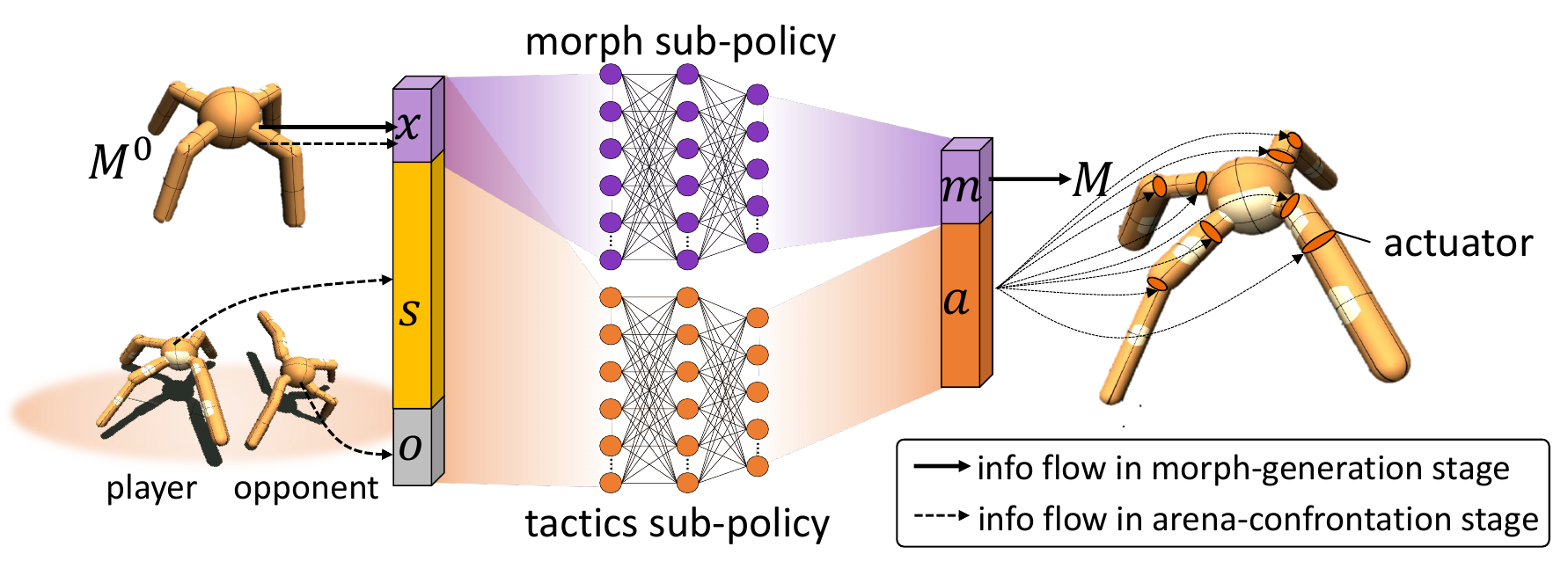}
    \caption{Information flow in morph and tactics co-evolution training. $x$ denotes initial parameters, which is a randomized vector; $s$ and $o$ are states of the agent and observation of the opponent, respectively; $m$ is generated encoded morph, and $a$ is generated actions applied to each actuator during the confrontation.}
    \label{fig:info flow}
\end{figure}

\subsubsection{Morph and Tactics Joint-optimization in Confrontation}
Thanks to the advancements in previous evolutionary algorithms~\cite{Ha2019,Wang2019,Yuan2022}, we now have a strong foundation for co-optimization. Unlike population-based bi-level optimization methods~\cite{Wang2019,Sims2023}, we take a more direct approach inspired by Transform2Act~\cite{Yuan2022}. Morph design parameters are generated from the original settings $x$ through the morph sub-policy network at the beginning of each episode and gradually converge with the growth of training generations (epochs).

We demonstrate self-practice co-evolution optimization in Figure~\ref{fig:info flow} and Algorithm~\ref{algo:algorithm}. We divide the process of adversarial co-evolution into two stages: morph-generation and arena-confrontation. First, we illustrate how to introduce morphological factors into an adversarial game in Figure~\ref{fig:info flow}. At the beginning of each game, we select policies for players and execute morphology generation to obtain morph pairs $M_{\alpha}$ and $ M_{\beta}$ to create an arena. The morph-generation stage is the first step of an episode, followed by the arena-confrontation stage.

More specifically, evolvable agents' morphology parameters $m$ are generated by policy $\pi(\theta)$ and then create a morph $M$ in the simulator. Note that for agents without morph evolution ability, we use fixed morphs and train only fighting tactics. Then combats start, and players concatenate ontological perception $s$, observation of opponent $o$, as well as morphology encoding $m$, generating actions through tactics sub-policy network. At the end of epochs, the latest policies of both players are collected and saved for later training. We sample data and update policies for two players by Proximal Policy Optimization(PPO), respectively, in every epoch.

\begin{algorithm}[t]
    \caption{Confrontation algorithm for co-evolving agents.}
    \label{algo:algorithm}
    \textbf{Input}: initial policies $\pi_{\alpha}^0$ and $\pi_{\beta}^0$, original morph $M_\alpha^0$ and $M_\beta^0$, opponent sampling factor $\delta$ \\
    \textbf{Parameter}: memory $\mathcal{M}$, policy pools $\mathcal{P}_{\alpha}$ and $\mathcal{P}_{\beta}$ \\
    \textbf{Output}: policies $\pi_{\alpha}$ and $\pi_{\beta}$
    \begin{algorithmic}[1] %[1] enables line numbers
    
    \STATE $\mathcal{P}_{\alpha} \leftarrow \emptyset$, $\mathcal{P}_{\beta} \leftarrow \emptyset$
    \STATE $\mathcal{P}_\alpha \leftarrow \mathcal{P}_\alpha \bigcup \left\{\pi_\alpha^0 \right\}, \mathcal{P}_\beta \leftarrow \mathcal{P}_\beta \bigcup \{\pi_\beta^0 \}$
    \WHILE{\textit{not reaching maximum generation}}
    \FOR{player $i$ in $\{\alpha, \beta \}$}
    \STATE $\mathcal{M} \leftarrow \emptyset$
    \STATE Define opponent player $j\ne i \; \mathrm{and} \; j\in \{\alpha, \beta \}$
    \STATE Sample policies $\pi_i,\pi_j$ from $\mathcal{P}_i, \mathcal{P}_j$
    \WHILE{\textit{episode not over}}
    \IF{\textit{at morph-generation stage}}
    \STATE Get $M_i, M_j$ by $m_i \sim \pi_i(\theta),m_j \sim \pi_j(\theta)$
    \STATE Create morph arena by $M_{i}, M_{j}$
    \STATE $r_{i} \leftarrow 0$; store $(r_{i}, m_{i})$ into $\mathcal{M}$
    \ELSIF{\textit{at arena-confrontation stage}}
    \STATE $a_{i} \sim \pi_{i}(\phi), a_{j} \sim \pi_{j}(\phi)$
    \STATE $s' \leftarrow \mathcal{T}(s, a_{i}, a_{j})$; $s \leftarrow s'$
    \STATE $r_{i} \leftarrow R_{i}(s,a_{i},a_{j})$
    \STATE Store $(r_{i}, a_{i}, m_{i}, s)$ into $\mathcal{M}$
    \ENDIF
    \ENDWHILE
    \STATE Update $\pi_i$ with PPO using data in $\mathcal{M}$
    \STATE $\mathcal{P}_i \leftarrow \mathcal{P}_i \bigcup \left\{\pi_i \right\}$
    \ENDFOR
    \ENDWHILE
    \STATE \textbf{return} $\pi_{\alpha}, \pi_{\beta}$
    \end{algorithmic}
\end{algorithm}

\section{Training Curriculum}
\label{training curriculum}

\subsection{Warming-up}
Training from scratch is extremely challenging, especially when the agent lacks any fundamental skills. To expedite the training process and progress toward learning confrontational skills, it is essential to initially train the agent in basic behaviors such as walking. This facilitates both agents to move rapidly toward each other, generating physical contact and confrontational interaction. Our designed warm-up rewards give the agent a better tendency to learn basic skills, including less energy consumption, higher speed, and correct moving directions. We generally train about 100 epochs to guide agents in basic skills.

Furthermore, appropriate guidance during the warm-up phase can help avoid some strange phenomena during competition training. For instance, training without warming up could lead to \texttt{spider} learning the behavior of jumping over an opponent to avoid physical confrontation in \texttt{run-to-goal}. These behaviors are undesirable since they are beyond the ability dimension of common players and are prone to premature strategy divergence. To avoid this, we guide agents to move forward as close to the ground as possible to prepare for the later physical confrontation.

\begin{figure}[t]
    \centering
    \subfigure[\texttt{run-to-goal}: agents are reset face-to-face and try to reach red lines behind the opponent as quick as possible.]{\includegraphics[width=0.236\textwidth]{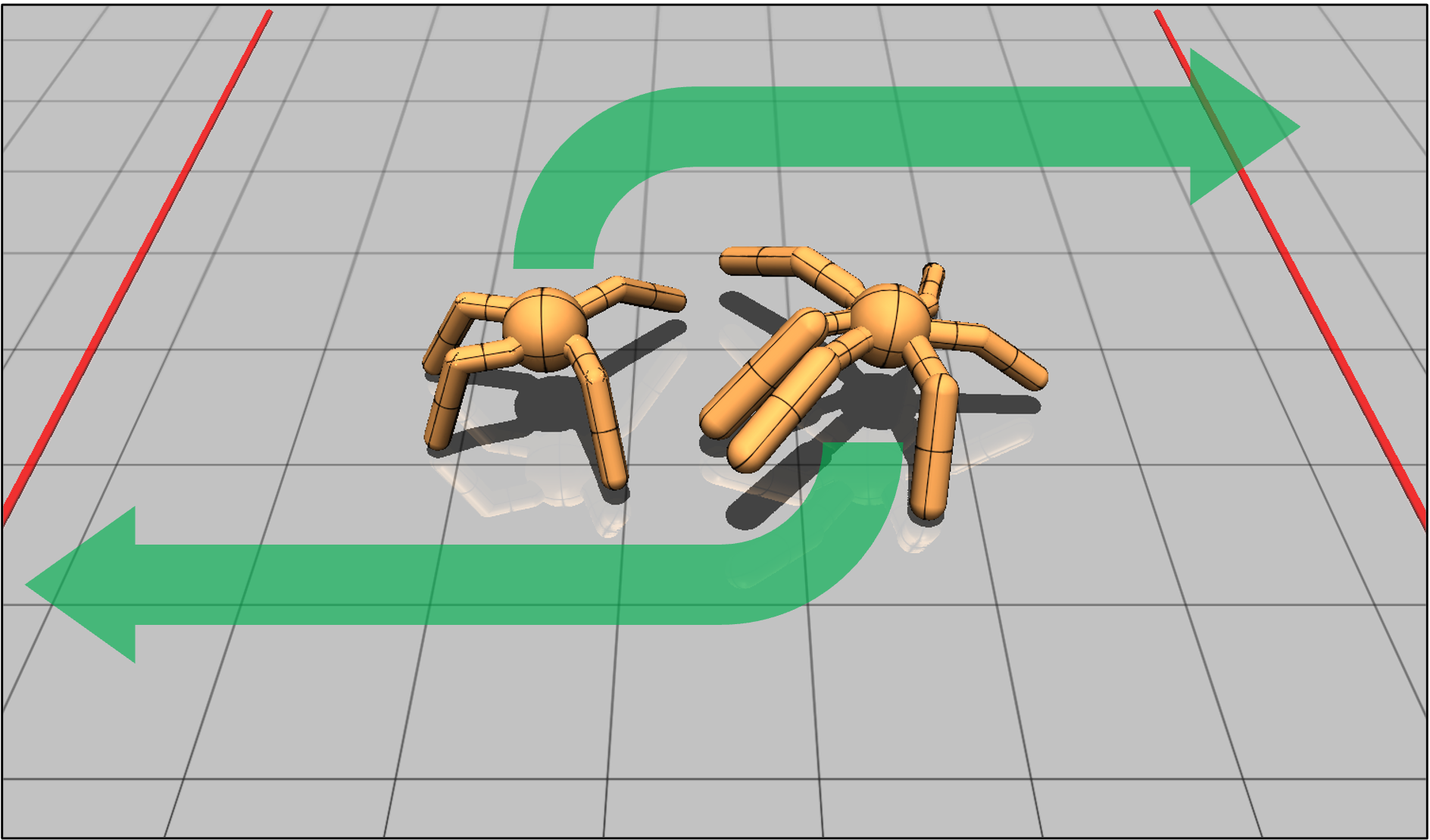}}
    \hfill
    \subfigure[\texttt{sumo}: agents fight on dohyo and try to push opponent out of the dohyo or knock opponent flat.]{\includegraphics[width=0.236\textwidth]{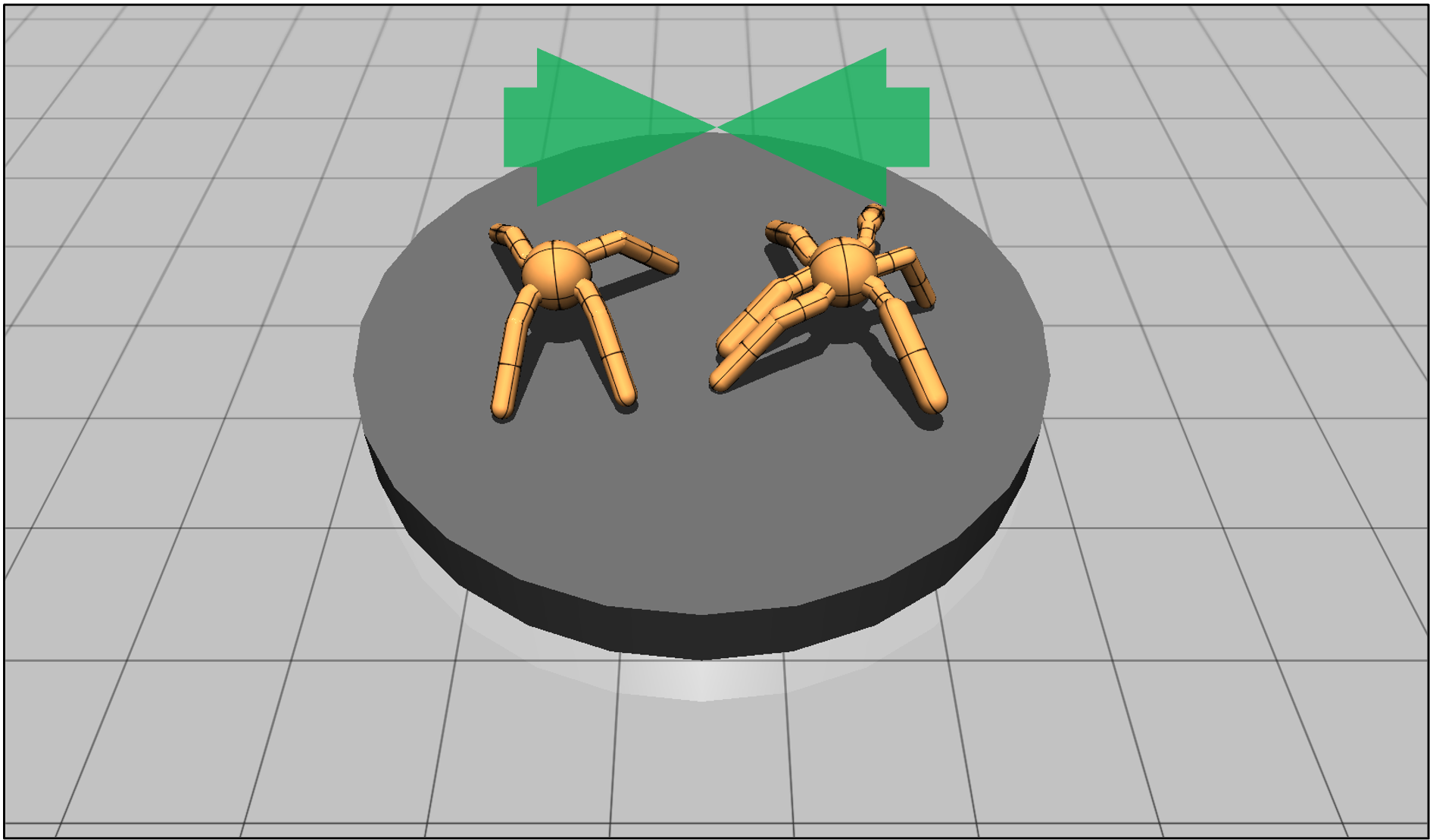}}
    \caption{Confrontation environments.}
    \label{fig:environments}
\end{figure}

\begin{figure*}[t]
    \centering
    \includegraphics[width=\textwidth]{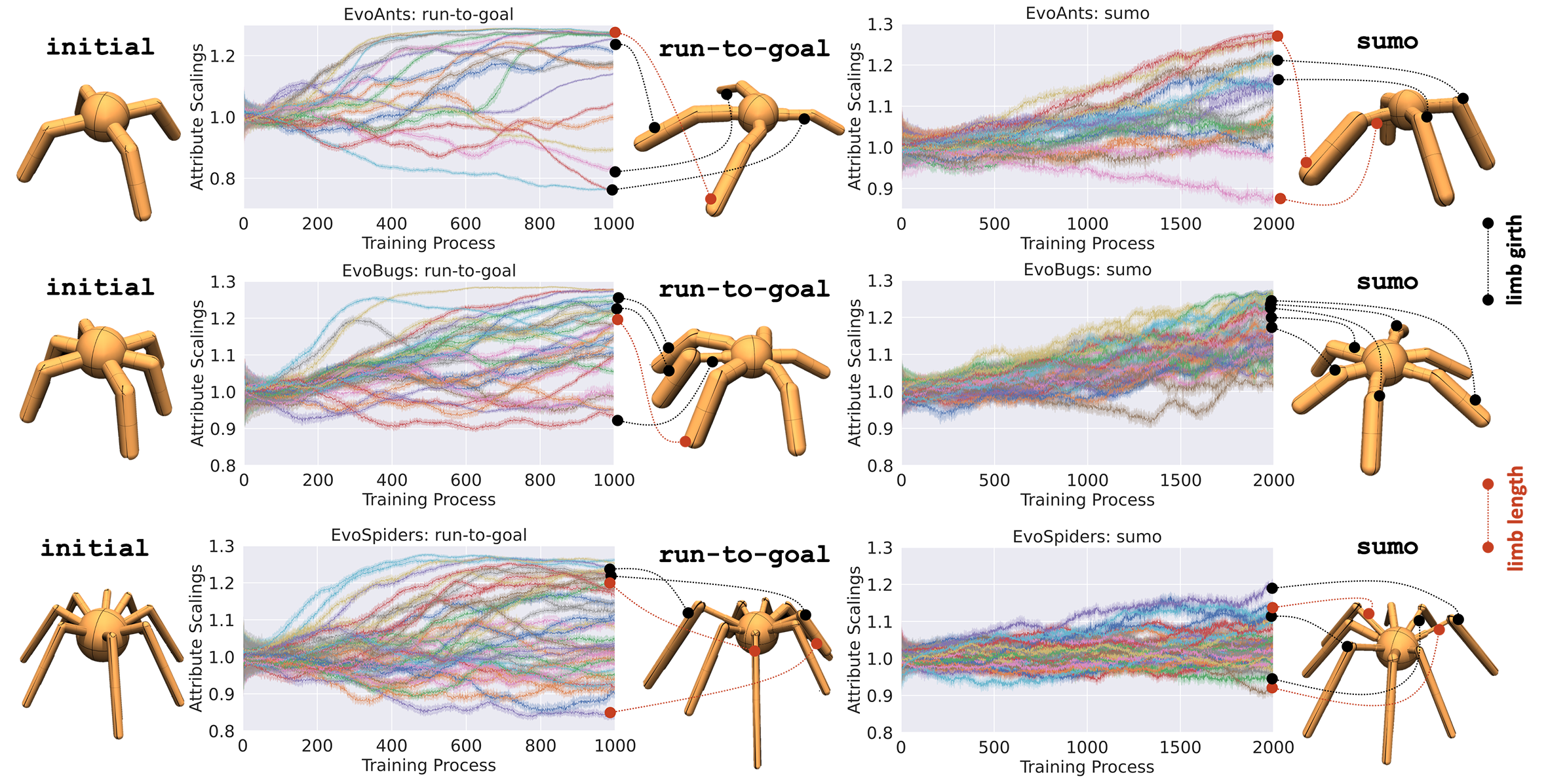}
    \caption{Morph parameter changes of evolvable agents in the training process and their final evolved morphologies. Each line represents a varying parameter. We mark with dashed lines the correspondence between the most noticeable changes in data and the physical profiles. The red dashed line corresponds to the length data of the limbs, while the black dashed line corresponds to the girth attributes of the limbs.}
    \label{fig:design changes}
\end{figure*}

\subsection{Rewards Annealing}
When the agents can move toward each other and make physical contact, confrontation training begins in earnest. At the end of every episode, we give winners a huge positive fighting reward and a negative one for losers. Nonetheless, fighting rewards in two-player games are too sparse for reinforcement learning, which makes training extremely challenging because the agent has to learn the correct actions with limited positive feedback. Training with only sparse rewards often leads to issues like signal delay, suboptimal problem~\cite{Riedmiller2018}, and high variance.

Therefore, we use both dense rewards and sparse rewards simultaneously to overcome exploration difficulty. The former encourages basic skills learning, and the latter provides stimulation for confrontation. For dense rewards design, we extend the reward settings from the warming-up phase to enhance the fundamental skills.

To balance dense rewards $R_d$ and sparse rewards $R_s$, we use an annealing factor $\kappa$ to make a trade-off. $\kappa$ varies depending on the current generation $t$ and predefined termination generation $T_t$:
\begin{align}
    R=\kappa R_{d} + (1-\kappa)R_{s},\; \kappa =\max \left ( \frac{T_t-t}{T_t},0  \right ) 
    \label{eq:annealing}
\end{align}
In the beginning, dense rewards dominate the training direction, then gradually reduce the influence until $\kappa$ declines to zero. After the termination generation $T_t$, only sparse rewards work. We generally set termination generation as the number of maximum training iterations, or half of it.

\section{Experiments}
\label{experiments}
\subsection{Environment Settings}
We implement CompetEvo in two physical contact adversarial environments: \texttt{run-to-goal} ~\cite{Bansal2018a} and \texttt{sumo}~\cite{MaruanAl-ShedivatTrapitBansalYuraBurdaIlyaSutskeverIgorMordatch2013}. In contrast to the previous environments, our morphological arenas permit each agent to evolve their morphs before the commencement of each game. Therefore, the morphological evolution strategy is contemporaneously and continuously updated in synchronization with confrontational events.

The goal of players in \texttt{run-to-goal} is to reach the red line behind the opponent, as illustrated in Figure~\ref{fig:environments}(a). Both agents will try to prevent any passing; meanwhile, they will also try to break through the adversary's defense and reach the red line. The first to reach the red line wins the game. Moving towards the goal can get dense rewards in every step. At the end of the game, we set a large sparse winning reward (+1000) for the winner and punishment (-1000) for the loser in \texttt{run-to-goal}. In \texttt{sumo} shown in Figure~\ref{fig:environments}(b), players try to push the opponent out of the arena or a knockout to win the game. Quickly pushing the opponent outside the arena, staying at the center of the arena, and moving towards the opponent, are possible ways to gain dense rewards. We set winning reward to +2000 and losing reward to -2000. Also, we penalize both sides (-1000) when a draw occurs to encourage confrontation; otherwise, agents might not learn aggressive behaviors.

During the training, we set maximum epochs to 1000 for \texttt{run-to-goal} and 2000 for \texttt{sumo}. Adam optimizer is used with a learning rate 0.0005. PPO clipping is 0.2, the discount factor is 0.995, and the generalized advantage estimate parameter is 0.95. 50,000 samples from 50 parallel rollouts are collected for one batch with mini-batches composed of 2,000 samples for PPO training. Termination generation $T_t$ is set to 1000 for both tasks.

% Our experiments are conducted on an Intel 13900K workstation with one NVIDIA RTX4090. The training duration primarily depends on the complexity of the physical contact during the data sampling process, which takes about 15 hours for \texttt{run-to-goal} and 28 hours for \texttt{sumo}.

To validate the performance of agents with distinct morphs, we set cross-antagonism for agents with the same training iterations and use their win rates over one hundred rounds to represent their abilities.

\begin{figure*}[t]
    \centering
    \includegraphics[width=\textwidth]{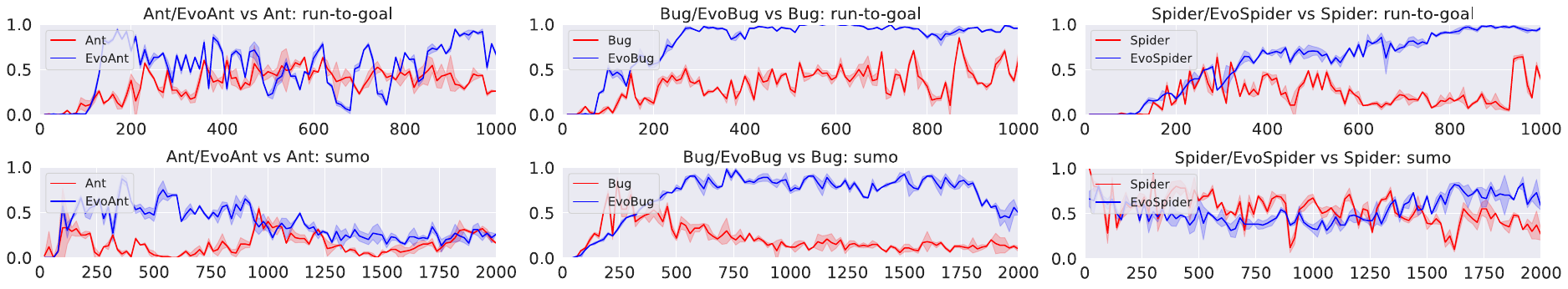}
    \caption{Win rates of evolvable and original agents in competition between symmetric species. The vertical axis denotes the win rate, and the horizontal axis denotes the iteration of the training process. The win rate shows a consistent increase throughout the training process. Particularly noteworthy is the significant improvement in the win rates of evolvable agents, indicated by the blue lines, in comparison to the original agents represented by the red lines.}
    \label{fig:winrate symmetric}
\end{figure*}

\begin{figure}
    \centering
    \subfigure[\texttt{run-to-goal}]{\includegraphics[width=0.235\textwidth]{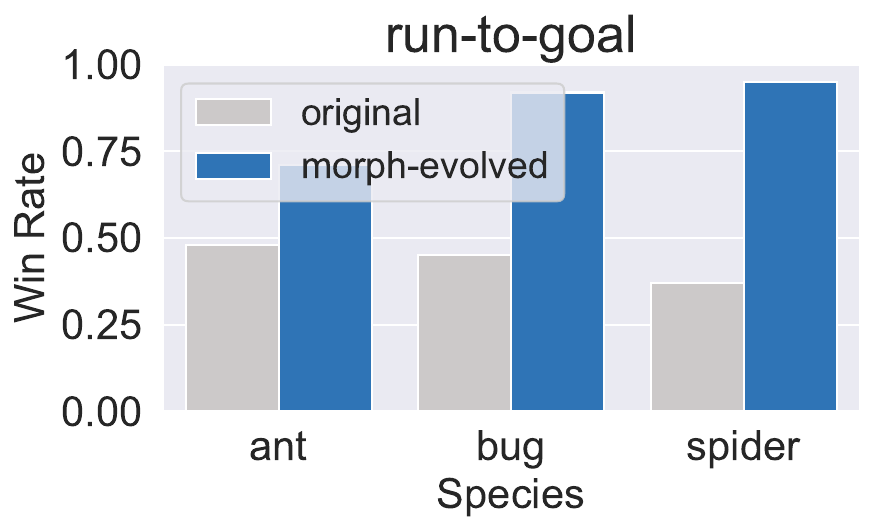}}
    \subfigure[\texttt{sumo}]{\includegraphics[width=0.235\textwidth]{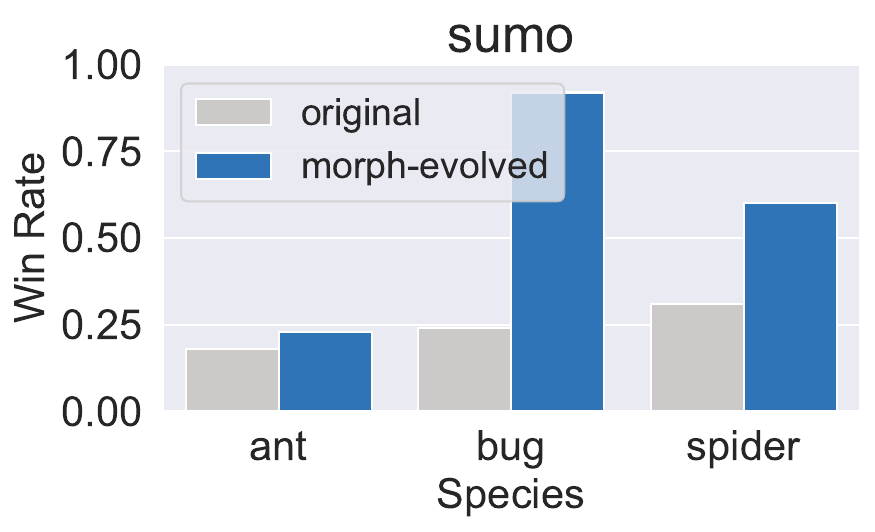}}
    \caption{Win rate comparison between evolvable agents and original agents when facing a different species.}
    \label{fig:winrate bar}
\end{figure}

\begin{figure*}[t]
    \subfigure[\texttt{run-to-goal}]{\includegraphics[width=\textwidth]{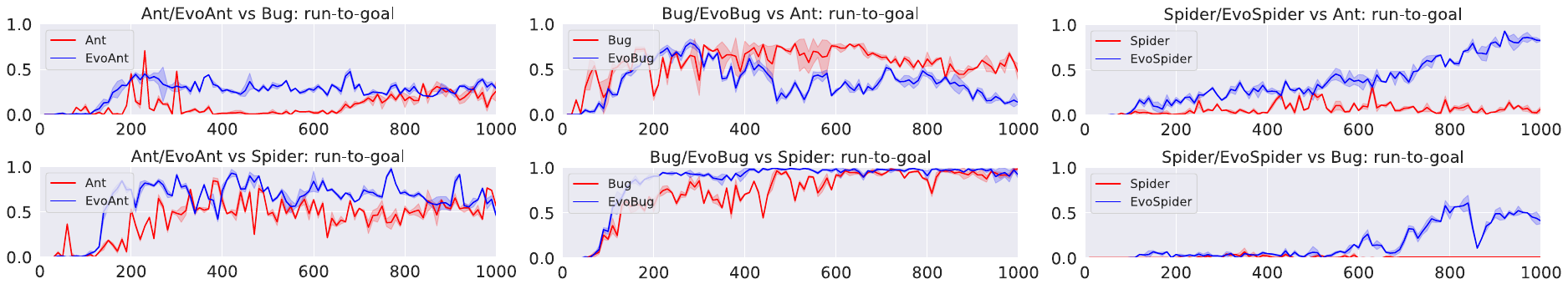}}
    \hfill
    \subfigure[\texttt{sumo}]{\includegraphics[width=\textwidth]{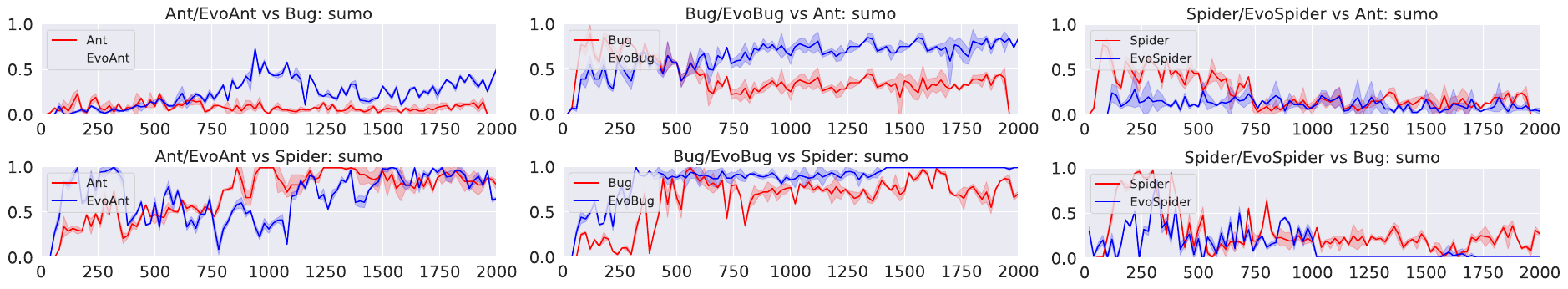}}
    \caption{Win rates of evolvable and original agents in competition between asymmetric species. The vertical axis denotes the win rate, and the horizontal axis denotes the iteration of the training process. }
    \label{fig:winrate asymmetric}
\end{figure*}

\subsection{Co-evolution in Self-Practice}
Evolvable agents simultaneously optimize their designs and fighting tactics through self-practice. During the training process, the design parameters' varying tendencies and the final morphs after converge are illustrated in Figure~\ref{fig:design changes}. 

The trend can be inferred by observing the variations. In \texttt{run-to-goal} competition, agents tend to evolve much more robust limbs towards the moving direction. This tendency helps to improve the stability of the heading and thus resist the interference of the opponent, as shown in the middle column in Figure~\ref{fig:design changes}. On the other hand, thighs tend to be thinner due to energy efficiency. This is because the lower legs play a more important role in propelling the agent than the thighs, and thighs degenerate due to energy saving. Right column in Figure~\ref{fig:design changes} illustrates that parameters growing in \texttt{sumo} are more concentrated and show the tendency to increase. Agents tend to develop brawny lower legs, and only very few limbs become thinner. This feature can be found on most legs because agents with stronger limbs can provide much more powerful impulsive force during confrontation to cope with threats from all directions.

\subsection{Effectiveness of Co-evolution}

% \begin{figure}
%     \centering
%     \subfigure[\texttt{run-to-goal}]{\includegraphics[width=0.235\textwidth]{pic/winrate_run-to-goal.pdf}}
%     \hfill
%     \subfigure[\texttt{sumo}]{\includegraphics[width=0.235\textwidth]{pic/winrate_sumo.pdf}}
%     \caption{Win rates of agents when facing all types of opponents. The numbers are the win rates of agents listed on the left axis.}
%     \label{fig:winrate}
% \end{figure}

As aforementioned, there are three species in our experiments: \texttt{ant}, \texttt{bug}, and \texttt{spider}, corresponding to their three original morphs and the three evolved morphs derived from them. To validate that morphological evolution can improve the capabilities of fighting, we first conduct experiments and set confrontations on symmetric species, between the original agents and their evolved versions. The expectation is that agents with evolved morphologies will achieve a higher win rate compared to their original design. Moreover, to further validate the effectiveness and generalization performance of evolved agents, we also conduct adversarial experiments among asymmetric species. We pit the original agent and the evolved version of one species against other species, comparing their respective winning rates.

\subsubsection{Comparison between Symmetric Species}
Results are shown in Figure~\ref{fig:winrate bar}. The win rates of morph-evolved agents are much higher than those of agents with original morphologies in all six scenarios, which indicates that our method can generate a more suitable design and strategy for confrontation. Besides, the fluctuations of the win rates varying over the training epochs are also shown in Figure~\ref{fig:winrate symmetric}. The blue lines represent the win rate of evolved agents fighting against original agents, while the red lines represent the win rates between two original agents. Nearly all evolvable agents consistently maintain a dominant position throughout the majority of the training process over two adversarial scenarios, which demonstrates the robustness and applicability of our method.

\subsubsection{Comparison between Asymmetric Species}
Further evidence can be found in asymmetric species. We select evolvable and original agents of the same species, engaging them in battles against agents from other species, and compare the resulting changes in their respective winning rates. According to results shown in Figure~\ref{fig:winrate asymmetric}, evolvable agents maintain higher win rates than those of their original morph in the majority of scenarios. Besides, in nearly half of the scenarios, the winning rates of evolvable agents get a promotion and eventually surpass those of their original morphs. There are also some scenarios where the effects are not pronounced. For example, in experiments of \texttt{spider} and \texttt{evo-spider} in \texttt{sumo} task, both of them get low win rates. This is because the spider agents have too many legs, requiring more precise cooperation between joints for movement, often leading to instability when facing external impacts. Additionally, the thin legs of spider agents inherently put them at a disadvantage in limb contact.

The main evidence consistently indicates that agents allowing for morphological evolution possess stronger capabilities. Nonetheless, this does not mean that evolved agents can defeat all physically weaker opponents of other species. There is only one failure case: \texttt{evo-bug} versus \texttt{ant} in \texttt{run-to-goal}. \texttt{ant} is much smaller, which helps it evade a frontal attack from a formidable opponent. It can hide under the \texttt{evo-bug}, lifting the opponent off the ground and carrying it to the goal.

Although competitions between different species are inherently unfair, our results reveal that in competitions between asymmetric species, employing a co-evolution approach enables naturally disadvantaged agents to grow up with more robust morphologies, thereby enhancing their winning rates.

\subsection{Typical Examples Illustrating the Roles of Morphological Evolution}

%\subsection{Emerging Behaviors by Morphological Evolution}

In this part, our focus is on highlighting the remarkable emergent behaviors resulting from morphological evolution. We conduct competitions by cross-confrontations among all six types of agents.

% and win rates are recorded in Figure~\ref{fig:winrate}. 

\begin{figure}[h]
    \centering
    \subfigure[Throwing out itself with robust limbs for higher speeds.]{\includegraphics[height=1.2in]{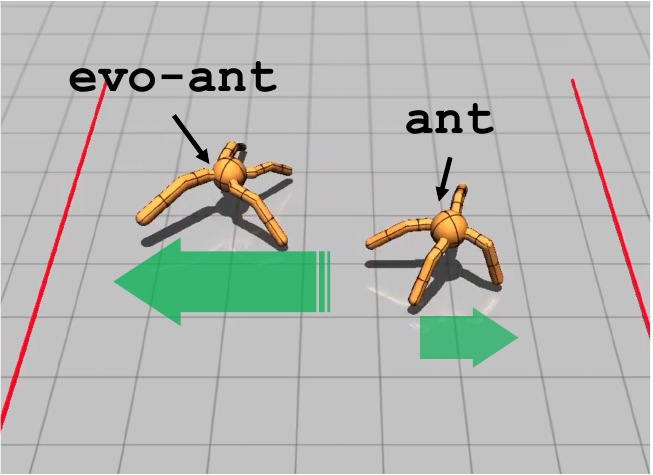}}
    \hfill
    \subfigure[Wrestling strategy to disrupt the rival's balance.]{\includegraphics[height=1.2in]{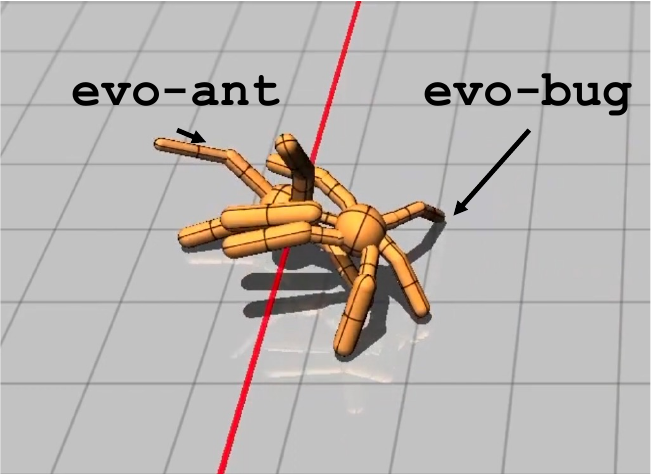}}
    \hfill
    \subfigure[Standing aided by limbs of varying lengths.]{\includegraphics[height=1.2in]{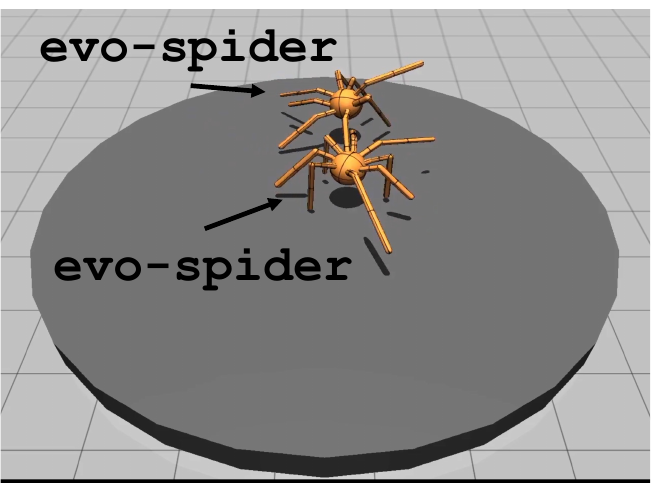}}
    \hfill
    \subfigure[Defending by expanding the limbs support angle.]{\includegraphics[height=1.2in]{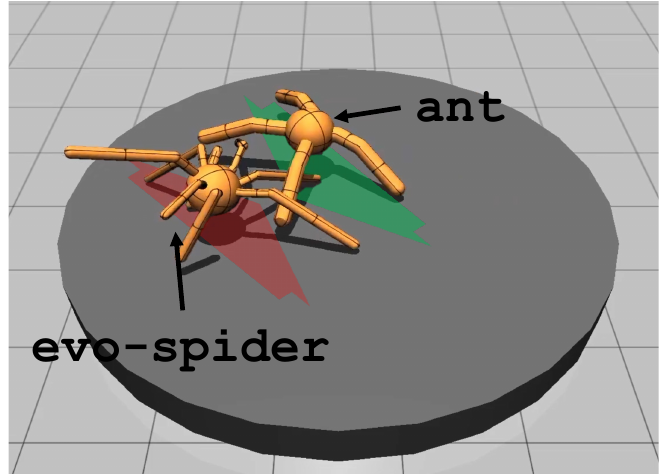}}
    \caption{Emerging behaviors related to morphological evolution.}
    \label{fig:emerging}
\end{figure}

The morphological evolution leads to interesting behaviors in the combat process. For example, evolvable agents tend to develop more robust limbs as they practice themselves, which gives them an unequivocal advantage over their original design. Here, we present four observed emergent behaviors resulting from the evolution of morphology: throwing, wrestling, standing, and defending, shown in Figure~\ref{fig:emerging}(a). In \texttt{run-to-goal}, we observe the behavior of throwing: agents evolve larger front legs, utilizing the inertia of the front legs to attempt to throw themselves forward, which is much faster than the original morphs. This phenomenon can be observed in all species of evolvable agents shown in Figure~\ref{fig:design changes}.

Besides, agents with morphological evolution exhibit limbs with greater strength, sufficient to employ techniques resembling wrestling moves to overturn opponents on the ground, as shown in Figure~\ref{fig:emerging}(b). The \texttt{evo-bug} develops three very strong legs, clamping onto one of the opponent's legs and then directly overturning them to the ground.

In task \texttt{sumo}, we also observed interesting behaviors corresponding to tactical strategies. In \texttt{evo-spider} versus \texttt{evo-spider}, spiders' numerous legs make it challenging to maintain coordination and balance to stand. Through morphological evolution, the spider evolved two to three relatively long legs among its many legs, forming a triangular support with the shorter legs. This arrangement prevents them from losing balance in most situations, shown in Figure~\ref{fig:emerging}(c).

Furthermore, morphological evolution gives rise to richer defensive tactics, illustrated in Figure~\ref{fig:emerging}(d). \texttt{evo-spider} evolves various lengths of legs. Longer side legs, as well as shorter front and back legs, are more advantageous for adopting a defensive posture. When facing opponents, \texttt{evo-spider} spread their legs to create a more stable point of force, resisting opponents' impacts, which can facilitate their defense tactics.

More results and interesting demonstrations can be found in the accompanying videos which are available at \href{https://competevo.github.io/}{https://competevo.github.io/}. Additionally, our environment files and related codes can be accessed from \href{https://github.com/KJaebye/competevo}{https://github.com/KJaebye/competevo}.

\section{Conclusion}
\label{conclusion}
In this work, we propose CompetEvo to introduce morphology evolution into multiagent competition tasks. The results show that co-evolving agent morphology and tactics can promote agents' combat ability. Our efforts represent a significant stride towards designing the most suitable agent for competition scenarios. More attempts can be made to evolve structural morphologies and create more meaningful scenarios like individual skeleton evolution in team games.

\section*{Acknowledgments}
This work was supported by the National Natural Science Foundation of China under Grant 62025304.

\section*{Ethical Statement}

There are no ethical issues.

%% The file named.bst is a bibliography-style file for BibTeX 0.99c
\bibliographystyle{named}
\bibliography{myref1}

\begin{thebibliography}{}

\bibitem[\protect\citeauthoryear{Al-Shedivat \bgroup \em et al.\egroup }{2018}]{MaruanAl-ShedivatTrapitBansalYuraBurdaIlyaSutskeverIgorMordatch2013}
Maruan Al-Shedivat, Trapit Bansal, Yura Burda, Ilya Sutskever, Igor Mordatch, and Pieter Abbeel.
\newblock {Continuous adaptation via meta-learning in nonstationary and competitive environments}.
\newblock {\em 6th International Conference on Learning Representations, ICLR 2018 - Conference Track Proceedings}, (March):1--21, 2018.

\bibitem[\protect\citeauthoryear{Bansal \bgroup \em et al.\egroup }{2018}]{Bansal2018a}
Trapit Bansal, Jakub Pachocki, Szymon Sidor, Ilya Sutskever, and Igor Mordatch.
\newblock {Emergent complexity via multi-agent competition}.
\newblock {\em 6th International Conference on Learning Representations, ICLR 2018 - Conference Track Proceedings}, 2018.

\bibitem[\protect\citeauthoryear{Cai \bgroup \em et al.\egroup }{2023}]{Cai}
Yishuai Cai, Shaowu Yang, Minglong Li, Xinglin Chen, Yunxin Mao, Xiaodong Yi, and Wenjing Yang.
\newblock {Task2Morph: Differentiable Task-Inspired Framework for Contact-Aware Robot Design}.
\newblock In {\em IEEE/RSJ International Conference on Intelligent Robots and Systems (IROS)}, pages 452--459, 2023.

\bibitem[\protect\citeauthoryear{Chen \bgroup \em et al.\egroup }{2023a}]{Chen}
Xinglin Chen, Da~Huang, Minglong Li, Yishuai Cai, Zhuoer Wen, Zhongxuan Cai, and Wenjing Yang.
\newblock {Evolving Physical Instinct for Morphology and Control Co-Adaption}.
\newblock In {\em IEEE/RSJ International Conference on Intelligent Robots and Systems (IROS)}, pages 6616--6623, 2023.

\bibitem[\protect\citeauthoryear{Chen \bgroup \em et al.\egroup }{2023b}]{Chen2023b}
Yang Chen, Yu~Luo, and Fuchun Sun.
\newblock {Cooperative distributed model predictive control for robot in-hand manipulation}.
\newblock {\em Robotic Intelligence and Automation}, 43(1):65--74, 2023.

\bibitem[\protect\citeauthoryear{Gleave \bgroup \em et al.\egroup }{2020}]{Gleave2020}
Adam Gleave, Michael Dennis, Cody Wild, Neel Kant, Sergey Levine, and Stuart Russell.
\newblock {Adversarial Policies: Attacking Deep Reinforcement Learning}.
\newblock {\em 8th International Conference on Learning Representations, ICLR 2020}, 2020.

\bibitem[\protect\citeauthoryear{Gupta \bgroup \em et al.\egroup }{2021}]{Gupta2021}
Agrim Gupta, Silvio Savarese, Surya Ganguli, and Li~Fei-Fei.
\newblock {Embodied intelligence via learning and evolution}.
\newblock {\em Nature Communications}, 12(1), 2021.

\bibitem[\protect\citeauthoryear{Ha}{2019}]{Ha2019}
David Ha.
\newblock {Reinforcement learning for improving agent design}.
\newblock {\em Artificial Life}, 25(4):352--365, 2019.

\bibitem[\protect\citeauthoryear{Haarnoja \bgroup \em et al.\egroup }{2023}]{Haarnoja2023}
Tuomas Haarnoja, Ben Moran, Guy Lever, Sandy~H. Huang, Dhruva Tirumala, Markus Wulfmeier, Jan Humplik, Saran Tunyasuvunakool, Noah~Y. Siegel, Roland Hafner, Michael Bloesch, Kristian Hartikainen, Arunkumar Byravan, Leonard Hasenclever, Yuval Tassa, Fereshteh Sadeghi, Nathan Batchelor, Federico Casarini, Stefano Saliceti, Charles Game, Neil Sreendra, Kushal Patel, Marlon Gwira, Andrea Huber, Nicole Hurley, Francesco Nori, Raia Hadsell, and Nicolas Heess.
\newblock {Learning Agile Soccer Skills for a Bipedal Robot with Deep Reinforcement Learning}.
\newblock {\em arXiv preprint arXiv:2304.13653}, 2023.

\bibitem[\protect\citeauthoryear{Hu \bgroup \em et al.\egroup }{2023}]{Hu2023}
Jiaheng Hu, Julian Whitman, and Howie Choset.
\newblock {GLSO: Grammar-guided Latent Space Optimization for Sample-efficient Robot Design Automation}.
\newblock {\em Proceedings of Machine Learning Research}, 205:1321--1331, 2023.

\bibitem[\protect\citeauthoryear{Huang \bgroup \em et al.\egroup }{2021}]{Huang2021c}
Kangyao Huang, Jingyu Chen, and John Oyekan.
\newblock {Decentralised aerial swarm for adaptive and energy efficient transport of unknown loads}.
\newblock {\em Swarm and Evolutionary Computation}, 67, 2021.

\bibitem[\protect\citeauthoryear{Huang \bgroup \em et al.\egroup }{2022}]{Huang2022c}
Kangyao Huang, Jingyu Chen, John Oyekan, and Xinyu Zhang.
\newblock {Bio-inspired Multi-agent Model and Optimization Strategy for Collaborative Aerial Transport}.
\newblock {\em Lecture Notes in Electrical Engineering}, 801 LNEE:591--598, 2022.

\bibitem[\protect\citeauthoryear{Huang \bgroup \em et al.\egroup }{2024}]{Huang2024}
Kangyao Huang, Di~Guo, Xinyu Zhang, Xiangyang Ji, and Huaping Liu.
\newblock {Stimulate the Potential of Robots via Competition}.
\newblock {\em arXiv preprint arXiv:2403.10487}, 2024.

\bibitem[\protect\citeauthoryear{Liu \bgroup \em et al.\egroup }{2019}]{Liu2019}
Siqi Liu, Guy Lever, Josh Merel, Saran Tunyasuvunakool, Nicolas Heess, and Thore Graepel.
\newblock {Emergent coordination through competition}.
\newblock {\em 7th International Conference on Learning Representations, ICLR 2019}, 2019.

\bibitem[\protect\citeauthoryear{Liu \bgroup \em et al.\egroup }{2022}]{Liu2022}
Siqi Liu, Guy Lever, Zhe Wang, Josh Merel, S.~M.Ali Eslami, Daniel Hennes, Wojciech~M. Czarnecki, Yuval Tassa, Shayegan Omidshafiei, Abbas Abdolmaleki, Noah~Y. Siegel, Leonard Hasenclever, Luke Marris, Saran Tunyasuvunakool, H.~Francis Song, Markus Wulfmeier, Paul Muller, Tuomas Haarnoja, Brendan Tracey, Karl Tuyls, Thore Graepel, and Nicolas Heess.
\newblock {From motor control to team play in simulated humanoid football}.
\newblock {\em Science Robotics}, 7(69), 2022.

\bibitem[\protect\citeauthoryear{Liu \bgroup \em et al.\egroup }{2023}]{liu2023morphology}
HP~Liu, D~Guo, FC~Sun, and X~Zhang.
\newblock Morphology-based embodied intelligence: Historical retrospect and research progress.
\newblock {\em Acta Autom. Sinica}, 49(6):1131--1154, 2023.

\bibitem[\protect\citeauthoryear{Riedmiller \bgroup \em et al.\egroup }{2018}]{Riedmiller2018}
Martin Riedmiller, Roland Hafner, Thomas Lampe, Michael Neunert, Jonas Degrave, Tom {Van De Wiele}, Volodymyr Mnih, Nicolas Heess, and Tobias Springenberg.
\newblock {Learning by playing - Solving sparse reward tasks from scratch}.
\newblock {\em 35th International Conference on Machine Learning, ICML 2018}, 10:6910--6919, 2018.

\bibitem[\protect\citeauthoryear{Schaff \bgroup \em et al.\egroup }{2019}]{Schaff2019}
Charles Schaff, David Yunis, Ayan Chakrabarti, and Matthew~R. Walter.
\newblock {Jointly learning to construct and control agents using deep reinforcement learning}.
\newblock {\em Proceedings - IEEE International Conference on Robotics and Automation}, 2019-May:9798--9805, 2019.

\bibitem[\protect\citeauthoryear{Silver \bgroup \em et al.\egroup }{2016}]{Silver2016}
David Silver, Aja Huang, Chris~J. Maddison, Arthur Guez, Laurent Sifre, George {Van Den Driessche}, Julian Schrittwieser, Ioannis Antonoglou, Veda Panneershelvam, Marc Lanctot, Sander Dieleman, Dominik Grewe, John Nham, Nal Kalchbrenner, Ilya Sutskever, Timothy Lillicrap, Madeleine Leach, Koray Kavukcuoglu, Thore Graepel, and Demis Hassabis.
\newblock {Mastering the game of Go with deep neural networks and tree search}.
\newblock {\em Nature}, 529(7587):484--489, 2016.

\bibitem[\protect\citeauthoryear{Silver \bgroup \em et al.\egroup }{2018}]{Silver2018}
David Silver, Thomas Hubert, Julian Schrittwieser, Ioannis Antonoglou, Matthew Lai, Arthur Guez, Marc Lanctot, Laurent Sifre, Dharshan Kumaran, Thore Graepel, Timothy Lillicrap, Karen Simonyan, and Demis Hassabis.
\newblock {A general reinforcement learning algorithm that masters chess, shogi, and Go through self-play}.
\newblock {\em Science}, 362(6419):1140--1144, 2018.

\bibitem[\protect\citeauthoryear{Sims}{2023}]{Sims2023}
Karl Sims.
\newblock {Evolving Virtual Creatures}.
\newblock {\em Seminal Graphics Papers: Pushing the Boundaries, Volume 2}, pages 699--706, 2023.

\bibitem[\protect\citeauthoryear{Vinyals \bgroup \em et al.\egroup }{2019}]{Vinyals2019a}
Oriol Vinyals, Igor Babuschkin, Wojciech~M. Czarnecki, Micha{\"{e}}l Mathieu, Andrew Dudzik, Junyoung Chung, David~H. Choi, Richard Powell, Timo Ewalds, Petko Georgiev, Junhyuk Oh, Dan Horgan, Manuel Kroiss, Ivo Danihelka, Aja Huang, Laurent Sifre, Trevor Cai, John~P. Agapiou, Max Jaderberg, Alexander~S. Vezhnevets, R{\'{e}}mi Leblond, Tobias Pohlen, Valentin Dalibard, David Budden, Yury Sulsky, James Molloy, Tom~L. Paine, Caglar Gulcehre, Ziyu Wang, Tobias Pfaff, Yuhuai Wu, Roman Ring, Dani Yogatama, Dario W{\"{u}}nsch, Katrina McKinney, Oliver Smith, Tom Schaul, Timothy Lillicrap, Koray Kavukcuoglu, Demis Hassabis, Chris Apps, and David Silver.
\newblock {Grandmaster level in StarCraft II using multi-agent reinforcement learning}.
\newblock {\em Nature}, 575(7782):350--354, 2019.

\bibitem[\protect\citeauthoryear{Wang \bgroup \em et al.\egroup }{2018}]{Wang2018}
Tingwu Wang, Renjie Liao, Jimmy Ba, and Sanja Fidler.
\newblock {Nervenet: Learning structured policy with graph neural networks}.
\newblock {\em 6th International Conference on Learning Representations, ICLR 2018 - Conference Track Proceedings}, 2018.

\bibitem[\protect\citeauthoryear{Wang \bgroup \em et al.\egroup }{2019}]{Wang2019}
Tingwu Wang, Yuhao Zhou, Sanja Fidler, and Jimmy Ba.
\newblock {Neural graph evolution: Towards efficient automatic robot design}.
\newblock {\em 7th International Conference on Learning Representations, ICLR 2019}, 2019.

\bibitem[\protect\citeauthoryear{Wang \bgroup \em et al.\egroup }{2023a}]{Wang2023c}
Minghao Wang, Ming Cong, Yu~Du, Dong Liu, and Xiaojing Tian.
\newblock {Multi-robot raster map fusion without initial relative position}.
\newblock {\em Robotic Intelligence and Automation}, 43(5):498--508, 2023.

\bibitem[\protect\citeauthoryear{Wang \bgroup \em et al.\egroup }{2023b}]{Wang2023}
Yuxing Wang, Shuang Wu, Tiantian Zhang, Yongzhe Chang, Haobo Fu, Qiang Fu, and Xueqian Wang.
\newblock {PreCo: Enhancing Generalization in Co-Design of Modular Soft Robots via Brain-Body Pre-Training}.
\newblock {\em Proceedings of Machine Learning Research}, 229:478--498, 2023.

\bibitem[\protect\citeauthoryear{Yuan \bgroup \em et al.\egroup }{2022}]{Yuan2022}
Ye~Yuan, Yuda Song, Zhengyi Luo, Wen Sun, and Kris~M. Kitani.
\newblock {Transform2Act: Learning a Transform-and-Control Policy for Efficient Agent Design}.
\newblock {\em ICLR 2022 - 10th International Conference on Learning Representations}, 2022.

\bibitem[\protect\citeauthoryear{Yuan \bgroup \em et al.\egroup }{2023}]{Yuan2023}
Lei Yuan, Ziqian Zhang, Ke~Xue, Hao Yin, Feng Chen, Cong Guan, Lihe Li, Chao Qian, and Yang Yu.
\newblock {Robust Multi-Agent Coordination via Evolutionary Generation of Auxiliary Adversarial Attackers}.
\newblock {\em Proceedings of the 37th AAAI Conference on Artificial Intelligence, AAAI 2023}, 37:11753--11762, 2023.

\bibitem[\protect\citeauthoryear{Zhang \bgroup \em et al.\egroup }{2022a}]{Zhang2022f}
Xinyu Zhang, Yuanhao Huang, Kangyao Huang, Xiaoyu Wang, Dafeng Jin, Huaping Liu, and Jun Li.
\newblock {A Multi-modal Deformable Land-air Robot for Complex Environments}.
\newblock {\em 2023 IEEE/ASME International Conference on Advanced Intelligent Mechatronics}, 2022.

\bibitem[\protect\citeauthoryear{Zhang \bgroup \em et al.\egroup }{2022b}]{Zhang2022g}
Xinyu Zhang, Yuanhao Huang, Kangyao Huang, Ziqi Zhao, Jingwei Li, Huaping Liu, and Jun Li.
\newblock {Coupled Modeling and Fusion Control for a Multi-modal Deformable Land-air Robot}.
\newblock {\em arXiv preprint arXiv:2211.04185}, 2022.

\end{thebibliography}

\end{document}